\documentclass[lineno,sn-mathphys,Numbered]{sn-jnl}

\usepackage{graphicx}%
\usepackage{multirow}%
\usepackage{amsmath,amssymb,amsfonts}%
\usepackage{amsthm}%
\usepackage{mathrsfs}%
\usepackage[title]{appendix}%
\usepackage{xcolor}%
\usepackage{textcomp}%
\usepackage{manyfoot}%
\usepackage{booktabs}%
\usepackage{algorithm}%
\usepackage{algorithmicx}%
\usepackage{algpseudocode}%
\usepackage{listings}%

\theoremstyle{thmstyleone}%
\theoremstyle{thmstyletwo}%

\theoremstyle{thmstylethree}%

\usepackage{hyperref} 
\usepackage{xurl}

\usepackage{bbm} 

\usepackage{multirow}  

\usepackage{pifont}
\newcommand{\cmark}{\ding{51}}%

\usepackage{colortbl}
\definecolor{Orange}{rgb}{0.9,0.5,0}
\definecolor{NavyBlue}{rgb}{0.1, 0.4, 0.8}
\definecolor{Magenta}{rgb}{0.8, 0.1, 0.6}
\definecolor{F7E0D5}{RGB}{247,224,213}
\colorlet{Light}{white!0!F7E0D5}

\usepackage[capitalize]{cleveref}  
\Crefname{section}{Sec.}{Sec.}
\Crefname{equation}{Eq.}{Eq.}
\Crefname{figure}{Fig.}{Fig.}
\Crefname{table}{Tab.}{Tab.}
\Crefname{appendix}{App.}{App.}

\begin{document}

\title[Article Title]{One Transformer for All Time Series: Representing and Training with  Time-Dependent Heterogeneous Tabular Data}


\author[1]{\fnm{Simone} \sur{Luetto}}\email{simone.luetto@prometeia.com}
\equalcont{These authors contributed equally to this work.}

\author[2,3]{\fnm{Fabrizio} \sur{Garuti}}\email{fabrizio.garuti@prometeia.com}
\equalcont{These authors contributed equally to this work.}

\author[3]{\fnm{Enver} \sur{Sangineto}}\email{enver.sangineto@unimore.it}

\author[2,4]{\fnm{Lorenzo} \sur{Forni}}\email{lorenzo.forni@prometeia.com}

\author[3,5]{\fnm{Rita} \sur{Cucchiara}}\email{rita.cucchiara@unimore.it}

\affil[1]{\orgname{Prometeia Spa}, \orgaddress{\city{Bologna}, \country{Italy}}}
\affil[2]{\orgname{Prometeia Associazione}, \orgaddress{\city{Bologna}, \country{Italy}}}
\affil[3]{\orgdiv{University of Modena and Reggio Emilia}, \orgname{AImageLab}, \orgaddress{\city{Modena}, \country{Italy}}}
\affil[4]{\orgdiv{University of Padova}, \orgname{Department of Economics}, \orgaddress{\city{Padova}, \country{Italy}}}
\affil[5]{\orgname{Istituto di Informatica e Telematica CNR}, \orgaddress{\city{Pisa}, \country{Italy}}}


\abstract{There is a recent growing interest in applying Deep Learning techniques to tabular data in order to
replicate the success of other Artificial Intelligence areas in this structured domain. 
Particularly interesting is the case in which tabular data have a time dependence, such as, for instance, financial transactions. 
However, the heterogeneity of the tabular values, in which categorical elements are mixed with numerical features, makes this adaptation difficult. 
In this paper we propose UniTTab,  a   Transformer based architecture whose goal is to uniformly  represent  heterogeneous time-dependent tabular data, in which both numerical and categorical features are described using continuous embedding vectors. Moreover, differently from common approaches, which use a combination of different loss functions for training with both numerical and categorical targets,  UniTTab
is  uniformly trained with a unique Masked Token pretext task. 
Finally, UniTTab can also represent  time series in which the individual row components  have a variable internal structure with a variable number of fields, which is a common situation in many application domains, such as in real world transactional data. Using extensive experiments with five datasets of variable size and complexity,  we empirically show that UniTTab consistently and significantly improves the prediction accuracy over several downstream tasks and with respect to both Deep Learning and more standard Machine Learning approaches.}




\maketitle

\section{Introduction}\label{sec.Introduction}

  Despite the success of Deep Learning methods in different areas of Artificial Intelligence (AI), such as, for instance, Natural Language Processing, Computer Vision, Audio Processing, Robotics, etc.,  the use of deep networks to represent tabular data is so far largely underexplored.
However, tabular data have a large application interest, since many public institutions or commercial/industrial  companies represent their structured knowledge using  datasets of ``tables'' \citep{Kotios}. For instance, bank data, clinical data, commercial data, etc., are often provided as a list of attributes (field names) and corresponding values (field values) for each represented entity (sample). 
As reported in \citep{Benjelloun-20},
over $65\%$ of the datasets in the Google Dataset Search platform contain tabular files in either CSV or XLS formats.
Particularly interesting is the case of financial transactions, which,  for instance, describe the sequence of (time dependent) transactions of a given bank client on their bank account.
However, most of the Machine Learning approaches for predictive tasks in these scenarios are often still based on pre-Deep Learning techniques, such as Gradient Boosted Decision Trees \citep{ye2024closerlookdeeplearning,DBLP:journals/ijdsa/BorisovBKK23,DBLP:conf/kdd/ChenG16,Kotios}.

One of the reasons  why Deep Learning is still underexplored with tabular data 
is   the lack of large scale publicly available datasets, often due 
to commercial or  privacy constraints (e.g., in the financial domain).
However, another important reason is the heterogeneity of these data, which are different from, e.g., texts or images and make their uniform representation and training challenging.
In fact, while an image is composed of a set of (numerical) pixel values and a textual sentence is a list of (categorical) words, tabular data are usually composed of heterogeneous attributes, in which numerical features (e.g., the transaction amount) are mixed with categorical features (e.g., the category of the transaction receiver or the transaction type). 
Thus, it is not clear how different types of features should be jointly represented, and what kind of objective function should be optimized during training to cover all the data types.
 Moreover,  a second source of heterogeneity comes from the necessity to represent 
 time sequences  composed of rows with a variable internal structure and a variable number of fields. For instance,  a ``Point Of Sale'' (POS)  transaction  includes a field describing the locality of the payment, which is not included in other transaction types made from {\em the same} bank account and included in {\em the same} time series (e.g., an ATM transaction).

In this paper, we propose a unified  architecture based on modern Transformers \citep{attention-is-all-you-need} to simultaneously and uniformly deal with all these representation problems. Specifically,    inspired from TabBERT \citep{TabBERT}, we use a hierarchical architecture to represent the time series dynamics. The hierarchy is composed of two  levels, the first of which represents a single tabular row (e.g., a specific transaction), while the second  represents a sequence of temporally dependent rows. This two-level hierarchy is a common solutions in, e.g., video processing, where each single frame is independently embedded  using a still image Transformer and the sequence of frames is then passed to a second Transformer \citep{DBLP:conf/iccv/Arnab0H0LS21}.  
We empirically show that this solution makes it possible to represent 
time series of more than one hundred rows.
However, differently from TabBERT and other similar hierarchical approaches, we introduce the following methodological novelties.

First, we modify the hierarchical network proposed in \citep{TabBERT} to uniformly and efficiently represent tabular rows with different {\em types}. Taking the financial transaction case as an example, a POS transaction corresponds to a table row with a set of fields different from an ATM transaction. We introduce a type-dependent embedding interface between the two levels of the hierarchy  which projects each row type into a fixed-dimension vector  fed to the second level. This interface extends the common look-up table of  initial token embeddings used in Transformer networks \citep{attention-is-all-you-need} and it
addresses a problem not yet considered in literature, i.e., how to deal with time series composed of structurally different types of consecutive rows. 

Second, inspired by the numerical representation adopted, for instance, in NeRFs \citep{DBLP:conf/eccv/MildenhallSTBRN20} for 3D synthesis, we propose to represent each numerical value as a feature vector obtained using  a set of sinusoidal functions at different frequencies. The motivation for this choice arises from the observation that   deep networks are biased towards learning low frequency functions \citep{DBLP:conf/icml/RahamanBADLHBC19}, while the scalar value of a specific tabular field undergoes to a high frequency variation. Thus, similarly to the coordinate embedding adopted in NeRFs, we smooth this variation using a set of sinusoidal transformations defined over a range of different frequencies. 
Note that this encoding is used to represent a specific {\em field value} and not to represent the  {\em field position} as with the standard Positional Encoding  used in Transformers \citep{attention-is-all-you-need}.

Third,   we adopt a unified objective function 
 for all the feature types based on the standard Masked Token pretext task \citep{devlin-etal-2019-bert}, which avoids the need to tune loss specific weights.
However, since, in our framework, the numerical features are {\em not} represented as discrete tokens, 
in order to use a Masked Token task for a numerical input, we propose to decouple the numerical feature representation from the target prediction used during the self-supervised training of the network. Specifically, inspired by BEiT \citep{BEiT}, where 
a discrete VAE  \citep{DALL-E} is used to quantize image patches and extract a discrete token for each of them, we similarly quantize the numerical features using a predefined set of bins, and then we use this quantization as the target token.
Note that the discretized value is the {\em target} label associated with a numerical value, but it is {\em not} used as input to the network.

Finally, we use a standard Label Smoothing \citep{DBLP:conf/cvpr/SzegedyVISW16} for the categorical feature tokens and we propose Neighborhood Label Smoothing for the numerical feature targets. The latter is based on the observation that numerical feature values, once discretized, still preserve a total order relation over the elements of the quantized vocabulary, thus, differently from pure categorical features, a small neighborhood of the original (numerical) value can be computed and used to restrict Label Smoothing on the most informative   value range.

We call our network  Unified Transformer for  Time-Dependent  Heterogeneous Tabular Data (UniTTab),  and we show that it consistently outperforms state-of-the-art approaches based on both Deep Learning and more standard Machine Learning techniques for time series of tabular data, often with a large margin. In our experiments, we use different common benchmarks and we additionally use a private dataset we collected in collaboration with a private financial institute. 
 It is a dataset composed of millions of real
 bank account transactions  spanning over 2 years, and it
represents a real life scenario, in which 
the transaction history of a given  account is  highly variable and different transactions of the same time series can have a different internal row structure.

\section{Related Work} \label{sec.Related}

Recent different works  focus on using  deep networks for tabular data \citep{ye2024closerlookdeeplearning} and time series \citep{trirat2024universaltimeseriesrepresentationlearning}. 
For instance, \citet{DBLP:conf/cikm/Lyu0ZG0TL22} propose an architecture which combines  different modules 
(meta-embeddings, automatic discretization
and aggregation) and can represent both numerical and categorical features. 
\citet{DBLP:journals/ijdsa/BorisovBKK23} use a distillation approach 
to map 
decision trees, trained on  heterogeneous tabular data,  onto homogeneous vectors which are fed to a deep network.
\citet{TabAConvBERT}  represent the time stamp as a sequence of different fields (e.g., the year, the month, etc.), each of which separately embedded and then summed together. We follow a similar paradigm when representing time, but we do not sum the corresponding embeddings.      
\citet{DBLP:journals/corr/abs-2012-06678} represent categorical features using an attribute specific embedding which is used as a prefix, concatenated with the actual field value.
\citet{DBLP:journals/corr/abs-2206-00664}  use a non-parametric representation of the training data, which reminds the use of external networks in Transformers  \citep{DBLP:conf/iclr/WuRHS22}.
\citet{beyazit2023an} represent a dataset of {\em numerical} tabular data using the Generalized Fourier Transform.
\citet{TabSyn} and \citet{suh2024timeautodiffcombiningautoencoderdiffusion} use a Variational AutoEncoder \citep{Kingma2014} to represent heterogeneous tabular data in a latent space and then they adopt a Diffusion
Model \citep{DDPMs} for data generation.
Other recent works use  Transformer architectures to represent tabular data 
\citep{DBLP:conf/nips/KossenBLGRG21,DBLP:conf/nips/GorishniyRKB21,DBLP:journals/corr/abs-2012-06678,DBLP:journals/corr/abs-2106-01342}. However,   these approaches still underperform standard, non Deep Learning based methods \citep{DBLP:journals/corr/abs-2012-06678,DBLP:conf/nips/GorishniyRKB21}, such as Gradient Boosted Decision Trees \citep{DBLP:conf/kdd/ChenG16}. 
Moreover, they usually represent numerical/categorical features in a relatively simple way, e.g., using a linear embedding layer or a quantization method, which we show
in \Cref{sec.Ablation} to be
 suboptimal. Finally, most of previous work 
dealing with
tabular data   does not model  the temporal {\em dynamics}: each row in the table is an individual sample. 
Conversely, in this paper  we focus on the more general case in which the  rows are concatenated into a sequence (a time series) representing the temporal evolution of the data, similarly to  frames in a video.

Another interesting line of work is based on directly or indirectly using natural language-based ``interfaces'' between the tabular data and a Transformer. For instance, in LUNA \citep{LUNA}, numerical values are represented as an atomic natural language string.
Similarly, \citet{DBLP:journals/corr/abs-2302-02041} represent numerical features using textual tokens.
Other works create a bridge between tabular data and Large Language Models (LLMs)
and use a zero/few-shot learning paradigm based on textual prompts.
For instance,
\citet{li2024unicluniversalcontrastivelearning}
use
an LLM-based encoder initialized with the pre-trained weights of  the textual encoder of CLIP
\citep{radford2021learning} to represent {\em numerical} time series of different domains.
\citet{DBLP:journals/pvldb/NarayanCOR22} transform numerical and categorical field values and attribute names in a natural-language prompt, and then they adopt a pre-trained GPT-3 language model \citep{DBLP:journals/corr/abs-2005-14165} as a few/zero shot learner. 
However,  the type of data used is relatively simple and should be described as attribute-value pairs (without any temporal dynamics).
In the context of synthetic tabular data generation, a similar strategy is adopted in \citep{DBLP:journals/corr/abs-2210-06280}, where syntactically correct natural language 
sentences are created using the feature names and the row values.
Similarly, \citet{Jiang-StructGPT-2022} develop a set of interfaces which are used to query a tabular dataset. The retrieved column names and values are then transformed in a textual prompt which is fed to a pre-trained LLM. 
However, apart from the lack of representation of temporal dynamics, there are other important problems in using a prompt to represent tabular data in natural language for an LLM. The first is that the LLM maximum context length limits the number of training  samples which can be fed to the LLM. Thus, when the training dataset is composed of, e.g., thousands or millions of rows, training a specialized Transformer usually leads to a performance largely superior to an in-context learning  paradigm \citep{DBLP:journals/pvldb/NarayanCOR22}.
The second is that, as observed by \citet{DBLP:journals/pvldb/NarayanCOR22}, tabular attributes and field values frequently contain domain-specific abbreviations or jargon terms not commonly found in textual documents, and, in these cases,  an LLM
lacks a strong semantic understanding of the input
because  these data  are out of its training distribution.

In this work, we train a specialized Transformer on large datasets of tabular data, and we use it to represent both the data heterogeneity and the  dynamics of the time series 
\citep{DBLP:journals/corr/abs-2202-07125}.
Specifically,  the work which is the closest to our proposal is TabBERT \citep{TabBERT}, in which the authors propose a hierarchical network to efficiently represent tabular time series (\Cref{sec.Introduction,sec.Preliminaries}).
However, our proposal differs from TabBERT in different aspects, the main of which are: (1) While in TabBERT numerical features are discretized in bins {\em when input to the network}, we propose a frequency-based embedding representation; (2) When training the network, we decouple the input and the output representations similarly to BEiT \citep{BEiT},
 and we  propose a Neighborhood-based Label Smoothing; (3) Our network architecture  can represent time series in which each row is an  ensemble of possibly different attributes, in this way dealing with real life complex datasets. In our experiments, we use TabBERT as the main baseline and we show that UniTTab  significantly outperforms TabBERT, as well as  all the other tested state-of-the-art models, in all the tested scenarios.

\begin{figure}[h]
    \centering
    \includegraphics[width=\columnwidth]{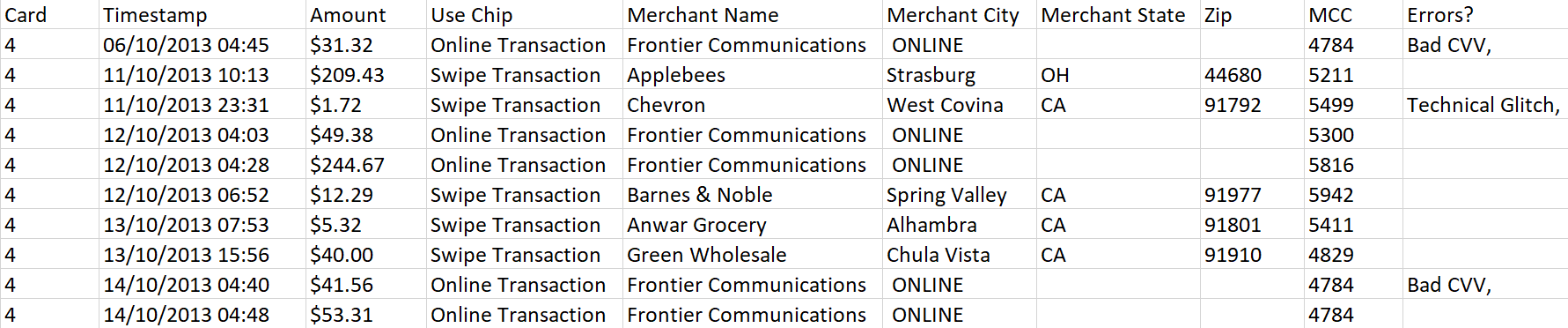}
    \caption{An example of time series taken from the Transaction Dataset \citep{TabBERT}. Each row is a bank transaction and it is composed of   $k = 10$  
    attributes.
    This sequence of $t = 10$ temporally consecutive transactions (rows) of the same client is a  time series.
    }
    \label{fig.transaction-dataset}
\end{figure}

\section{Preliminaries}
\label{sec.Preliminaries}

{\bf Problem Statement.}
Tabular data are represented as a set of attributes (field names) $A = \{ a_1, ..., a_k \}$, where each $a_j \in A$ can be either  categorical or numerical, and a set of table rows $\pmb{r}_1, ..., \pmb{r}_N$, which specify a value for each field: $\pmb{r}_i = [v_1, ..., v_k]$. 
If $a_j$ is numerical, then $v_j \in \mathbb{R}$, otherwise $v_j \in V_j$, where $V_j$ is an unordered attribute-specific vocabulary of categories.
A time series is a (variable length) sequence of rows $\pmb{s} = [\pmb{r}_1, ..., \pmb{r}_t]$  which are related to each other by a temporal dynamics.
For instance, in financial transactional data, $\pmb{s}$ can represent  the last $t$ bank account transactions of a given  client (see \Cref{fig.transaction-dataset}). 
When adopting a Deep Learning method,
a common paradigm is to    use a large dataset of time series to pre-train a  network with self-supervised learning, and then  fine-tune the model for a specific downstream task using task-specific labeled data and a possible smaller (supervised) dataset.

In this paper, we further generalize the previous scenario introducing time series composed of different row types. As mentioned in \Cref{sec.Introduction}, this generalization is particularly useful in real life datasets, in which, for instance, a transaction time series is composed of different transaction types (e.g., POS type, ATM type, etc.). Formally, we describe this situation using a function which associates each row in $\pmb{s}$ to a predefined set of row types: $type(\pmb{r}_i) = h \in T = \{ 1, ..., n \}$ and using a type-dependent set of attributes $A_h = \{ a_1, ..., a_{k_h} \}$ to specify the fields of $\pmb{r}_i$. Note that the cardinality of the attributes ($k_h$) varies depending on $h$.
Moreover, we assume that $type(\pmb{r}_i)$ is always defined for each $\pmb{r}_i \in \pmb{s}$: for example, each transaction in a time series of a bank account can be of only one type (e.g., POS, or ATM, etc.).

\begin{figure*}[h]
    \centering
    \includegraphics[width=\linewidth]{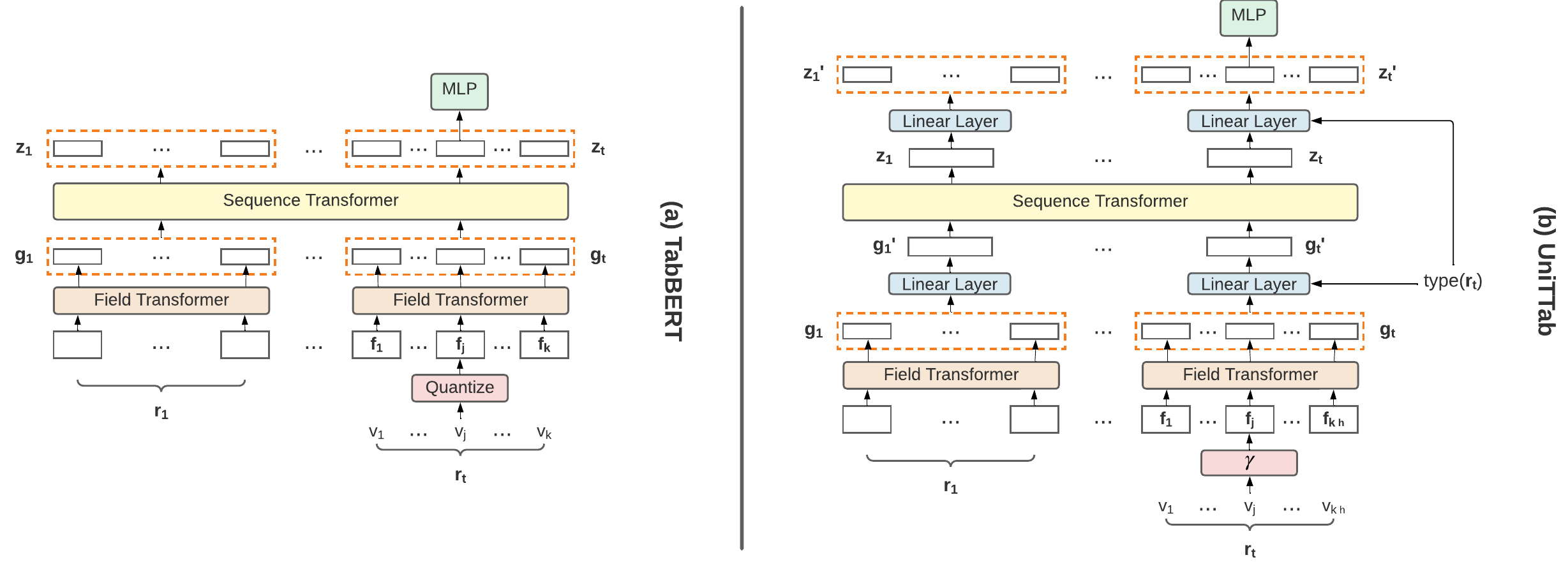}
    \caption{A schematic comparison between the architectures of TabBERT (a) and UniTTab (b).
    In both figures, $v_j$ is a numerical vale. Note that in (b) the number of attributes of each row ($k_h$) is variable.
    }
    \label{fig.architecture}
\vskip -0.1in
\end{figure*}

{\bf TabBERT architecture.} 
TabBERT is a hierarchical architecture composed of two different Transformers, trained end-to-end (\Cref{fig.architecture} (a)). The first Transformer (called ``Field Transformer'') takes as input 
the $k$ field values of
a single table row $\pmb{r}_i$. Note that $k$ is constant for all the rows, as TabBERT implicitly assumes that there is only one row type (in our notation: $|T| = n = 1$). 
Numerical features are discretized using an attribute specific set of bins. In this way, both numerical and categorical features can be associated to a specific discrete token. The tokens are transformed in embedding vectors using a standard  look-up table of learned embeddings \citep{attention-is-all-you-need}. In \Cref{fig.architecture} (a), this is indicated as a set of field embeddings $\pmb{f}_1, ..., \pmb{f}_k$.
The Field Transformer transforms these vectors in $k$ final embeddings 
of dimension $d$,
which are concatened in a single vector $\pmb{g}$ of dimensions $d \cdot k$. Then,
$\pmb{g}$ is  fed to the second Transformer
(``Sequence Transformer''), jointly with the representations of all the other rows in the input 
time series $\pmb{s}$. 
Note that the dimension of each $\pmb{g}$ should be constant because 
$\pmb{g}$ is a fixed-size  initial embedding vector for the second Transformer.
The Sequence Transformer outputs 
 a sequence of $t$ final embedding vectors $\pmb{z}_1, ..., \pmb{z}_t$. 
 Finally, each $\pmb{z}_i$ is 
 split in $k$ vectors, on top of which a shallow MLP is used to output a posterior distribution over the attribute-specific vocabulary. This makes it possible to apply a Masked Token pretext task \citep{devlin-etal-2019-bert} during pre-training, in which a few  tokens are randomly masked and the network is asked to predict the masked tokens.

\section{Method}
\label{sec.Method}

In this section we present  UniTTab, showing the architecture of the network, the way in which heterogeneous features are represented and the uniform pre-training strategy.

{\bf Row-type dependent embedding.} 
We first extend  the hierarchical architecture of TabBERT  to deal with a variable number of row types ($ n > 1$). The main problem we need to solve is that $k_h$ depends on $type(\pmb{r}_i)$ for each $\pmb{r}_i \in \pmb{s}$, while the dimension of $\pmb{g}$ should be fixed (\Cref{sec.Preliminaries}). We solve this problem using a linear projection layer (\Cref{fig.architecture} (b)) which takes $h = type(\pmb{r}_i)$ as input and transforms 
$\pmb{g} \in \mathbb{R}^{d \cdot k_h}$ in $\pmb{g}' \in \mathbb{R}^{m}$, where $m$ is fixed and the transformation depends on a row-type specific linear matrix $W_h$:

\begin{equation}
\label{eq.row-specific-transform}
\pmb{g}' = W_h \pmb{g} \: \: \: \: \: \:  (W_h \in \mathbb{R}^{m \times (d \cdot k_h)}).
\end{equation}

The set of learnable projection matrices $W_1, ..., W_n$, one for each row type, constitutes a look-up table of embeddings for the initial layer of the second Transformer, and naturally extends the common initial embedding look-up table used in Transformer networks. Analogously, each  final row embedding $\pmb{z}$ (\Cref{sec.Preliminaries}) is transformed in $\pmb{z}' \in \mathbb{R}^{d \cdot k_h}$ using a specific weight matrix $S_h$ ($S_h \in \mathbb{R}^{(d \cdot k_h) \times m}$)
before being fed to the final MLP (\Cref{fig.architecture} (b)).

{\bf Feature representation.} 
We represent each categorical feature using a standard linear embedding based on its attribute-specific vocabulary. However, for numerical features, we extract a frequency-based representation as follows. Let $v$ be a scalar value corresponding to a numerical attribute. Similarly to \citep{DBLP:conf/eccv/MildenhallSTBRN20}, we transform $v$ using:

\begin{equation}
    \gamma(v) = (\sin(2^0 \pi v), \cos(2^0 \pi v), ..., \sin(2^{L-1} \pi v), \cos(2^{L-1} \pi v) ),
\end{equation}

\noindent
where $L = 8$ is the number of sine/cosine pairs used (see \Cref{sec.Ablation}). The vector $ \gamma(v)$ so obtained is then fed to a linear (learnable) layer whose output is the initial embedding vector for $v$.
Finally, similarly to \citep{TabAConvBERT},
we represent the numerical value of a time stamp attribute (e.g, measured in minutes)   using a combination of different time fields (i.e, the year, the month, the day, and, if necessary, the hour). Each such basic value is represented as a categorical feature  (e.g., with 12 elements for the month, etc.). In preliminary experiments, we also tried to represent each basic time field of a date as a numerical attribute (i.e. by means of our frequency-based representation), but this led to slightly worse results, presumably because the periodicity of some time series (e.g., in transactional data) can  better be represented by the network as a categorical value. Note, that, differently from \citep{TabAConvBERT}, we do not add the corresponding field embeddings but treat them as additional fields simply by increasing the number of attributes of each single row.

{\bf Training.} 
During the unsupervised pre-train stage, we train UniTTab  using {\em only}
a Masked Token pretext task. Specifically, for an input sample $\pmb{s} = [\pmb{r}_1, ..., \pmb{r}_t]$, we 
randomly replace a   field value $v \in \pmb{r}_i$ with the special symbol \texttt{[MASK]}. We use a standard  replacement probability value
 $p_f = 0.15$ \citep{devlin-etal-2019-bert,TabBERT}.
Moreover, with probability $p_r = 0.1$,
we also mask {\em all} the values in a  row $\pmb{r}_i$, while,
for the fields representing the time stamp, they are 
 always either jointly masked or jointly unmasked. We call these additional masking strategies ``Row masking'' and ``Time stamp masking'', inspired by the block masking  of adjacent image patches used in BEiT \citep{BEiT}, and we use them to make the pretext task more challenging for the network. 

Note that we can replace (the initial embedding vector of) $v$ with \texttt{[MASK]} independently on whether $v$ is numerical or categorical.
However,  the problem is what the network should predict in correspondence of $v$ if this is not an element of a discrete vocabulary. A possible solution could be to adopt a regression loss function and directly ask to the network to reconstruct the original numerical value $v$. The disadvantage of this hybrid solution is that we would need two different loss functions: one for the categorical features (e.g., the Cross Entropy), and another for the numerical values (e.g., an MSE loss), which then need to be suitably weighted. Conversely, we propose a different solution: inspired by BEiT, we quantize $v$ and we use its categorical representation as the target label. Specifically,
if $a_j$ is a numerical attribute, we define a vocabulary of bin values $V_j = \{b_1, ..., b_q \}$ 
spanning the whole range of possible values for $a_j$. Then, when a given value $v$ for this attribute is masked, we quantize $v$ ($b = quantize(v)$) and we use $b \in V_j$ as a pseudo label for $v$, which is used as the token to be predicted. Note that $b$ is {\em not} input to the network. 

When we compute the loss function, we use Label Smoothing \citep{DBLP:conf/cvpr/SzegedyVISW16}, which replaces the one-hot vector representation of a categorical label $v \in V_j$ with a
vector of smoothed probability values
 $\pmb{p}(v)$. 
 In more detail, if $V_j = \{ v_1, ..., v_{q_j} \}$ is the vocabulary corresponding to $a_j$, and  $q_j$ is its cardinality, then a ground truth value $v \in V_j$ is associated with the vector $\pmb{p}(v)$ defined as follows:
\begin{equation}
\label{eq.label-smoothing-categ}
    [\pmb{p}(v)]_l = 
    \left\{ 
\begin{array}{ll}
1 - \epsilon,  & \mbox{if  $l = v$} \\
\frac{\epsilon}{q_j - 1}  & \mbox{otherwise,} \\
\end{array}
\right.
\end{equation}
\noindent
where $[\pmb{p}(v)]_l$ is the $l$-th element of $\pmb{p}(v)$ and $\epsilon$ a small constant.
Furthermore, in case $a_j$ is a numerical attribute, 
 we propose Neighborhood Label Smoothing, in which  $b = quantize(v)$ is smoothed using only a small neighborhood centered in $b$. Specifically, we use the range $R = \{ b - 5, ..., b + 5 \}$, and we replace \Cref{eq.label-smoothing-categ}
 with:
\begin{equation}
\label{eq.label-smoothing-numer}
    [\pmb{p}(v)]_l = 
    \left\{ 
\begin{array}{ll}
1 - \epsilon,  & \mbox{if  $l = b$} \\
\frac{\epsilon}{10}  & \mbox{if  $l \in R, l \neq b$} \\
0  & \mbox{otherwise.} \\
\end{array}
\right.
\end{equation}
 Note that this is possible because, if $a_j$ is numerical, then $V_j$, obtained with quantization, is an ordered set.

Finally, if $\pmb{s}'$  is the perturbed version of $\pmb{s}$, in which some random field values have been masked as explained above, then
our  Masked Token pretext task can be formulated as minimizing the following Cross Entropy loss:
\begin{equation}
\label{eq.masked-token-loss}
    \min_{\theta} - \sum_{v \in \pmb{s}} \mathbbm{1}_{Masked(v)} \pmb{p}(v) \log p_{\theta} (quantize(v) | \pmb{s}'),
\end{equation}
\noindent
where: $\theta$ are the parameters of the network, 
$p_{\theta} (y | \pmb{x})$ is the probability of the network to predict $y$ given the sequence $\pmb{x}$ as input,
$\mathbbm{1}_{Masked(v)}$ is 1 only when $v$ was masked (otherwise is 0), $quantize(v) = v$ if $v$ is categorical, and,
with a slight abuse of notation, $v \in \pmb{s}$ indicates a generic  field value of one of the rows in $\pmb{s}$.

\section{Experiments}
\label{sec.Results}

In this section, we evaluate UniTTab using different datasets and downstream tasks. We use the same architecture for all the datasets, with the only difference being the number of fields ($k$) and the length ($t$) of the time series, which depend on the specific dataset and the adopted evaluation protocol. Specifically, 
independently of the dataset, we always use only one self-attention layer  with 8 heads in the Field Transformer, and 12 self-attention layers  with 12 heads each in the Sequence Transformer. 
For a fair comparison, the total number of parameters of UniTTab was kept approximately the same as in TabBERT (see \Cref{sec:hyperparameters} for more details).
We use five different-size datasets of  time series of heterogeneous tabular data: the Pollution Dataset \citep{PollutionDataset},  the Transaction Dataset \citep{TabBERT} (both adopted by TabBERT), the  PKDD’99 Financial Dataset \citep{Berka-Dataset},
the Age2 dataset \citep{Age2},
and our Real Bank Account Transaction Dataset (in short, RBAT Dataset). In the latter, each time series  is composed of  three different row types  (i.e., $n = |T| = 3$, see \Cref{sec.Preliminaries}). The other four datasets have only one row type ($n = |T| = 1$), thus, in all the experiments but  those  on  the RBAT Dataset, we use   only one projection matrix in \Cref{eq.row-specific-transform}. 
 All the five datasets are composed of tabular data with both numerical and categorical fields (see \Cref{sec:Dataset-statistics} for more details).
In all the experiments,  the Deep Learning based models are first pre-trained using self-supervision (\Cref{sec.Method}) and then evaluated using a dataset specific (supervised) downstream task, while standard Machine Learning methods are directly trained on the downstream task training set.
Moreover, to enable a fair comparison with other methods tested on
 the Pollution Dataset and  the Transaction Dataset,  we follow  the protocol defined in \citep{TabBERT} for the downstream task definition and the training/testing partitions. However, since
 the stride used in \citep{TabBERT} to extract the time series from the Pollution Dataset  leads to a partial information leak  between the training and the testing data of the downstream task,
 we additionally use a different training/testing partition for this dataset, which we call ``our partition'', while we use the term ``original partition'' to refer to the initial split   adopted in \citep{TabBERT} and in other works
(see \Cref{sec:setting} for more details). Finally, for the other three datasets, we define 
the details of the corresponding downstream tasks  and the training/testing partitions in \Cref{sec:setting}.

\begin{table*}[t]
\vspace{2mm}
\setlength{\tabcolsep}{2.0pt}
\footnotesize{
    \caption{Ablation study using the Pollution Dataset (our partition) and a  subset of the Transaction Dataset.}
    \label{tab.ablation}
    \centering
    \begin{tabular}{ccccccccll}
    \toprule
     & Quantized  & Combina.   & Frequency   & Regression   &  CE loss   & Standard  & Neighbor  & 
     {\bf Pollution} & {\bf Transaction} \\
      & numerical   & time & based num.   & loss  &   (only)  & Label  & Label & {\bf Dataset} & {\bf Dataset} \\    
    & features  & stamp  & features     &      &           & Smoothing & Smoothing 
     & (RMSE $\downarrow$) & (F1-score $\uparrow$) \\
    \midrule
   1 &  \cmark &   &   &   &  \cmark  &   &   &  34.30 & 0.829 \\
   2 &    \cmark   & \cmark   &   &   &  \cmark  &   &   &   32.23  & 0.840 \\
   3 &    &     & \cmark & \cmark &    &   &   &   32.47  & 0.844 \\
   4 &    &     & \cmark &   &  \cmark &   &   &   31.52  &  0.846 \\
   5 &     & \cmark   & \cmark &   &  \cmark &   &   &   29.63  & 0.847 \\
   6 &     & \cmark   & \cmark &   &  \cmark &  \cmark &   &   29.47  & 0.848 \\
   7 &     & \cmark   & \cmark &   &  \cmark & \cmark & \cmark &  \bf 29.05 &  \bf 0.850 \\
    \bottomrule
    \end{tabular}
    }
\end{table*}

\begin{table}[t]
\vspace{2mm}
\setlength{\tabcolsep}{4.0pt}
\small
    \caption{An analysis of different masking strategies using the Pollution Dataset (our partition) and a  subset of the Transaction Dataset.}
    \label{tab.masking}
    \centering
    \begin{tabular}{ccccll}
    \toprule
  & Row masking  & Row masking  & Time stamp  & {\bf Pollution Dataset} & {\bf Transaction Dataset} \\
   &  $p_r = 0.05$ &  $p_r = 0.1$ &  masking &   (RMSE $\downarrow$) & (F1-score $\uparrow$) \\
    \midrule
    1 &  &   &   &  29.05 & 0.850  \\
    2 &     \cmark   &   &   &  29.01 & 0.851  \\
    3 &   & \cmark &   &  28.91 & 0.854 \\
    4 &   &   & \cmark &   28.82 & 0.855 \\
    5 &   & \cmark & \cmark &  \bf 27.99  & \bf 0.858 \\
    \bottomrule
    \end{tabular}
\end{table}    

\begin{table}[t]
\vspace{2mm}
\setlength{\tabcolsep}{4.0pt}
\small
    \caption{The influence of the number of frequency function pairs ($L$)  using the Pollution Dataset (our partition) and a  subset of the Transaction Dataset.}
    \label{tab.L-ablation}
    \centering
    \begin{tabular}{cccll}
    \toprule
       &    {\bf Pollution Dataset} & {\bf Transaction Dataset} \\
    $L$    &    (RMSE $\downarrow$) &    (F1-score $\uparrow$) \\
    \midrule
     4  &  28.63 & 0.828  \\
     6  &  28.38   & 0.830  \\
     8  &   \bf 27.99 & \bf 0.858  \\
     10  &  28.11 & 0.808  \\
     12  &  28.20  & 0.804 \\
    \bottomrule
    \end{tabular}
\end{table}

\subsection{Ablation study}
\label{sec.Ablation}

In this section, we analyze the contribution of each component of our method. We use the Pollution Dataset 
 \citep{PollutionDataset} 
 and a random subset of the Transaction Dataset \citep{TabBERT} (400K time series, see \Cref{sec:setting} for more details).
The former dataset is composed of time series where each row contains $k=11$ fields, 10 of which are numerical attributes and one is categorical (\Cref{sec:Dataset-statistics}). Hence, this dataset is particularly suitable for investigating the influence of different numerical feature representations (\Cref{sec.Method}). The downstream task is a regression task, and the performance is measured using RMSE (the lower the better). In the Transaction Dataset, the rows of the time series are composed of 2 numerical  and 8 categorical attributes, and its associated downstream task is a binary classification task. In this case,  we use the F1 score as the performance metric (the higher the better). Both datasets contain a time stamp attribute, which we represent using  4 time attributes as explained in \Cref{sec.Method}.
For both datasets, following \citep{TabBERT}, after
self-supervised pre-training,  
 the  final embeddings of the Sequence Transformer ($\pmb{z}_1, ..., \pmb{z}_t$) are used as input to an LSTM which is (separately) trained to solve  a supervised regression or binary classification task
 using the corresponding task labels.
Specifically, we first train 
UniTTab using \Cref{eq.masked-token-loss}  and then we train
an LSTM (using the same LSTM architecture adopted in \citep{TabBERT}) on top of the 
final embeddings of our Sequence Transformer.
In \Cref{sec.Main-Results} we also show the results obtained by directly fine-tuning our model, without using an LSTM, as well as comparative results obtained using the full Transaction Dataset and   both the original and our  partition of the Pollution Dataset.

\Cref{tab.ablation} analyses the impact of the main components of our method. The first row is the baseline, which corresponds to our implementation of TabBERT, starting from its publicly available code and keeping fixed all the main hyperparameters (e.g., the number of layers  of the two Transformers, the number of heads, the emdedding size, etc.).
In this baseline, the numerical features are quantized into discrete bins and treated as categorical {\em when input to the network}. 
 Note that there is no Row or Time stamp masking nor Label Smoothing.
The  the Cross Entropy loss is the only loss used because the tokens of the quantized numerical features can be masked and predicted just like categorical tokens.
The second row of the table shows the improvement obtained when the time stamp is split in 4  different fields, which corresponds to a significant performance improvement in both datasets and tasks. In the third row, we replace the discrete representation of the numerical features with our frequency-based representation (\Cref{sec.Method}). In this case, for each masked numerical field value $v$, we use a regression function, which consists in predicting the original scalar value of $v$ (before the frequency-based embedding). We use the squared difference  between the predicted and the ground truth value as the loss function (MSE loss) for these numerical features, which is summed to the Cross Entropy loss computed with the categorical features.
Note that, as mentioned in \Cref{sec.Method}, one of the problems when using two different loss functions is the necessity to weight their relative importance. In this ablation experiment, we  set the MSE loss weight using a simple heuristic in which we compute the average MSE loss value and the average Cross Entropy loss value and we impose these two values to be equivalent using a relative weight (more details in \Cref{sec:hyperparameters}).
Comparing this row with the baseline (first row), the improvement obtained when using a  frequency-based numerical representation jointly with a regression loss is comparable with the introduction of the combinatorial time stamp.
In  row 4,  
the regression loss is replaced by the Cross Entropy loss, which is applied to numerical features using pseudo target labels  (see \Cref{eq.masked-token-loss}).   
Comparing this row with the baseline, we obtain the largest relative improvement, which shows that 
using frequency-based numerical representations jointly with pseudo labels for the Cross Entropy 
can significantly boost  the network performance.
In row 5, the combinatorial time stamp is added to the setting of row 4, bringing an additional benefit. 
Finally, row 6 and row 7  show the results corresponding to the introduction of standard Label Smoothing and Neighborhood Label Smoothing, respectively.

In \Cref{tab.masking}, we analyze the impact of using Row masking and Time stamp masking, where the first strategy is evaluated with two different selection probability values $p_r$. The best result corresponds to using both Row masking (with probability 
$p_r =  0.1$) 
and Time stamp masking. 
In \Cref{tab.L-ablation} we empirically evaluate the influence of the number of frequency function pairs $L$ (\Cref{sec.Method}) used for our frequency-based numerical feature representation. In the rest of this paper, we use the best value ($L = 8$) for all the other datasets and tasks.

Finally, we postpone the ablation of the Row-type dependent embedding  (\Cref{sec.Method}) to \Cref{sec.Main-Results}, where we introduce our RBAT Dataset composed of
variable row types. 
In the same section, we sow the impact of different amount of pre-training data on the downstream task performance.


\subsection{Main results}
\label{sec.Main-Results}

In this section, we compare UniTTab with different state-of-the-art approaches using different datasets and downstream tasks. 

{\bf Pollution prediction task.} This is the regression task used in \Cref{sec.Ablation} (and presented with  more details in \Cref{sec:setting}).
 We show results based on two main paradigms: 
based on LSTM training  and on directly fine-tuning the pre-trained model. 
The former is based on the protocol proposed by \citet{TabBERT}, and trains a separate LSTM using 
the
vectors $\pmb{z}_1, ..., \pmb{z}_t$  as input and the labels of the downstream task as the supervision (see \Cref{sec.Ablation}).
However, the latter, proposed here,  is likely a much more natural choice, and it is coherent with most of the recent AI literature, where the backbone network, after pre-training, is directly fine-tuned for a specific downstream task without training a separate  network. Specifically, when we fine-tune either UniTTab or TabBERT, we include a  \texttt{[CLS]} token in the input sequence of the  Sequence Transformer and we  use the corresponding $\pmb{z}_{\texttt{[MASK]}}$ final embedding vector as input to a final linear layer dedicated to the specific task (and trained from scratch).
In case of LUNA \citep{LUNA}, we report the results we obtained using its public code  and only the LSTM training paradigm (adopted also in \citep{LUNA}), because fine-tuning this method requires non-trivial modifications of its architecture.

To enable a comparison with results published in previous work, in the experiments
of \Cref{tab.pollution-TabBERT-split} we strictly follow \citep{TabBERT} and we use the original partition of the Pollution Dataset (see \Cref{sec.Results} and \Cref{sec:setting}) jointly with a time series length $t = 10$. Conversely, 
in \Cref{tab.pollution-our-split} we use our partition (where $t = 10$ but also the time series extraction stride is 10, see \Cref{sec:setting} for more details) and 
we repeat all the experiments 5 times with different random seeds.
Note that the  random seed is used both to   initialize the models' parameters and to randomly split the time series in training and testing splits.
Finally, in \Cref{tab.pollution-t-50} we extend these experiments to longer time series with $t=50$ 
(\Cref{sec:setting}).
In all cases (\Cref{tab.pollution-our-split,tab.pollution-TabBERT-split,tab.pollution-t-50}), UniTTab outperforms both TabBERT and LUNA by a large margin, and, as expected, direct fine-tuning significantly improves over the separate LSTM training.

In the same tables,  we also report the results obtained using 
both XGBoost \citep{DBLP:conf/kdd/ChenG16}
and CatBoost \citep{CatBoost}, which are the state-of-the-art non Deep Learning based methods for tabular data
(\Cref{sec.Introduction,sec.Related}).  Specifically, 
since both XGBoost and CatBoost cannot directly work on (variable-length) time series, we used a standard library \citep{CHRIST201872} to extract features from a time series. These “engineered” features include field-specific statistics (e.g., the mean or the variance for numerical features and the mode for the categorical ones, etc.), autocorrelation with different “lags”, the number of local minima and maxima, etc. Moreover, the features are automatically pre-selected using a Filter feature selection method on a validation set extracted and separated  from the training data.
Finally, we used grid search on the validation set 
to set the optimal values of the XGBoost  and the CatBoost hyperparameters (e.g., the number of trees, the max depth of each tree, etc.). 
After feature selection and hyperparameter tuning, the validation set is merged with the training set and XGBoost/CatBoost are re-trained 
using the same data adopted for training the LSTMs or fine-tuning the UniTTab/TabBERT models on the supervised downstream task.
We will use this procedure (task and dataset dependent feature selection and hyperparameter tuning) for XGBoost   CatBoost in all the other
 downstream tasks of this paper.
 Additionally, we use Vector AutoRegression (VAR) \citep{RePEc:spr:sprbok:978-3-540-27752-1}, which is a common (non Deep Learning based) autoregressive model for multivariate time series, where we transform categorical features in numerical by assigning a numerical value to each element in the vocubaluary $V_j$.
In case of VAR, the field values of all the rows of the time series are directly input to the model without intermediate engineered features.
The results  in \Cref{tab.pollution-our-split,tab.pollution-TabBERT-split,tab.pollution-t-50} show that UniTTab largely  outperforms
``standard'' Machine Learning approaches, and the gap is particularly significant with longer sequences (\Cref{tab.pollution-t-50}), where the hierarchical Transformer architecture of both UniTTab and TabBERT can likely better represent long inter-row dependencies.



\begin{table}[h]
\caption{Pollution prediction task (original partition). 
\textsuperscript{\dag}~Our reproduction.
\textsuperscript{\ddag}~Results  reported in the corresponding  paper.
}
\label{tab.pollution-TabBERT-split}
\begin{tabular}{@{}lll@{}}
\toprule
Downstream task training & Model &   RMSE  \\ \midrule
\multirow{3}{*}{Fine-tuning} 
& TabBERT \dag    & 24.10 \\ 
&  UniTTab (ours) &    {\bf 20.05}    \\ \midrule
\multirow{3}{*}{LSTM} 
& TabBERT \citep{TabBERT}  \ddag   & 32.80 \\ 
& LUNA  \dag    &  37.73 \\ 
&  UniTTab (ours) &    {\bf 25.42}    \\ \midrule
\multirow{2}{*}{Training from scratch} 
&  XGBoost \textsuperscript{\dag} &    34.14    \\ 
&  CatBoost \textsuperscript{\dag} &    34.15    \\
& VAR \textsuperscript{\dag} &  83.83 \\
\bottomrule
\end{tabular}
\end{table}

\begin{table}[h]
\caption{Pollution prediction task (our partition): average and standard deviation results obtained with 5 random seeds. 
\textsuperscript{\dag}~Our reproduction  using the official code.}
\label{tab.pollution-our-split}
\begin{tabular}{@{}lll@{}}
\toprule
Downstream task training & Model &   RMSE  \\ \midrule
\multirow{2}{*}{Fine-tuning} 
& TabBERT  \textsuperscript{\dag}    & 31.41 $(\pm 1.74)$ \\
&  UniTTab (ours) &    {\bf 25.37 } $(\pm 1.59)$  \\ \midrule
\multirow{3}{*}{LSTM} 
& TabBERT   \textsuperscript{\dag}   & 37.20 $(\pm 1.67)$ \\ 
& LUNA    \textsuperscript{\dag}   & 45.17 $(\pm 0.62)$ \\ 
&  UniTTab (ours) &    {\bf 30.88} $(\pm 1.70)$   \\ \midrule
\multirow{3}{*}{Training from scratch} 
&  XGBoost \textsuperscript{\dag} &    48.89  $(\pm 0.73)$  \\ 
&  CatBoost \textsuperscript{\dag} &    47.09 $(\pm 0.89)$   \\ 
& VAR \textsuperscript{\dag} &  84.05 $(\pm 1.71)$  \\

\bottomrule
\end{tabular}
\end{table}

\begin{table}[h]
\caption{Pollution prediction task (our partition) with longer time series ($t=50$). Average and
standard deviation results obtained with 5 random seeds. † Our
reproduction using the official code.
}
\label{tab.pollution-t-50}
\begin{tabular}{@{}lll@{}}
\toprule
Downstream task training & Model &   RMSE  \\ \midrule
\multirow{3}{*}{Fine-tuning} 
& TabBERT \dag    & 30.95 $(\pm 0.53)$ \\ 
&  UniTTab (ours) &    {\bf 25.11} $(\pm 0.48)$   \\ \midrule
\multirow{3}{*}{LSTM} 
& TabBERT \cite{TabBERT}  \ddag   & 36.67 $(\pm 0.81)$ \\ 
& LUNA  \dag    &  43.93 $(\pm 2.83)$ \\ 
&  UniTTab (ours) &    {\bf 28.17} $(\pm 0.51)$     \\ \midrule
\multirow{3}{*}{Training from scratch} 
&  XGBoost \textsuperscript{\dag} &    56.37 $(\pm 0.92)$     \\ 
&  CatBoost \textsuperscript{\dag} &    56.26 $(\pm 0.99)$     \\
&  VAR \textsuperscript{\dag} &    83.37 $(\pm 0.95)$    \\  \bottomrule
\end{tabular}
\end{table}


\begin{table*}[h]
\caption{Fraud detection task ($t=10$).
\textsuperscript{\dag}~Our reproduction.
\textsuperscript{\ddag}~Results  reported in the corresponding  paper.
}
\label{tab.Fraud-detection}
\begin{center}
\small{
\begin{tabular}{@{}lll@{}}
\toprule
Downstream task training  & Model &   F1 score  \\ 
 \midrule
\multirow{3}{*}{Fine-tuning} 
& TabBERT \textsuperscript{\dag}     & 0.910  \\ 
& TabAConvBERT \citep{TabAConvBERT}  \ddag    & 0.896  \\
&  UniTTab (ours) &    {\bf 0.915}    \\ \midrule
\multirow{3}{*}{LSTM} 
& TabBERT \ddag \citep{TabBERT}    & 0.860 \\ 
& LUNA \ddag \citep{LUNA}    & 0.862  \\ 
&  UniTTab (ours) &    {\bf 0.914}     \\ \midrule
Training from &  XGBoost \textsuperscript{\dag} &    0.779    \\ 
scratch &  CatBoost \textsuperscript{\dag} &    0.499   \\
& VAR \textsuperscript{\dag} &   0.495   \\

\bottomrule
\end{tabular}
}
\end{center}
\end{table*}

\begin{table*}[h]
\caption{Fraud detection task ($t=50$).
\textsuperscript{\dag}~Our reproduction.
}
\label{tab.Fraud-detection-t-50}
\begin{center}
\small{
\begin{tabular}{@{}lll@{}}
\toprule
Downstream   & Model &   F1 score \\ 
task training &  &      \\ \midrule
\multirow{3}{*}{Fine-tuning} 
& TabBERT \textsuperscript{\dag}     & 0.777  \\ 
&  UniTTab (ours) &    {\bf 0.812}    \\ \midrule
\multirow{3}{*}{LSTM} 
& TabBERT \dag \cite{TabBERT}    & 0.883  \\ 
& LUNA \dag \cite{LUNA}    & 0.884\\ 
&  UniTTab (ours) &    {\bf 0.931}    \\ \midrule
Training from &  XGBoost \textsuperscript{\dag} &    0.349   \\ 
scratch &  CatBoost \textsuperscript{\dag} &    0.337   \\
&  VAR \textsuperscript{\dag} &    0.497    \\ \bottomrule
\end{tabular}
}
\end{center}
\end{table*}

{\bf Fraud detection task.}
This is the classification task used in \Cref{sec.Ablation} (and presented in more detail in  \Cref{sec:setting}). Differently from \Cref{sec.Ablation},
in the experiments of this section we use the full dataset for pre-training ($\sim 4,9$M time series), which leads to a higher absolute performance in the  results reported in \Cref{tab.Fraud-detection}.
Specifically, we use the original time series length $t=10$ in \Cref{tab.Fraud-detection} and we extend the experiments to $t=50$ in \Cref{tab.Fraud-detection-t-50}.
These tables
shows that UniTTab 
outperforms all the compared  methods, independently on whether an LSTM is used or not and  the gain is particularly significant with longer time series (\Cref{tab.Fraud-detection-t-50}).
Similarly to the Pollution prediction task, in \Cref{tab.Fraud-detection,tab.Fraud-detection-t-50} we also report the results obtained using VAR, XGBoost and CatBoost,  with  task and dataset specific  features and  hyperparameters for XGBoost and CatBoost (see above). Also in the 
Fraud detection   task, UniTTab significantly outperforms all the non Deep Learning based methods,
and the improvement is particularly significative with longer time series (\Cref{tab.Fraud-detection-t-50}).

{\bf Loan default prediction task.}
For this task, we use the  PKDD’99 Financial Dataset \citep{Berka-Dataset},  
which is smaller than the other datasets (see \Cref{sec:setting,sec:Dataset-statistics} for more details), thus we 
report the average results obtained with 5 random splits. 
The dataset consists of the complete (real) transaction history of several bank customers, the average length of which is 232.
To increase the number of time series available for pre-training, we set a maximum length value $t_{max}$, and we use 
 time series with variable  length  $t$, where $t \leq t_{max}$.
 Specifically, 
 for each  client, assuming that 
$t_{all}$ is  her/his total number of  transactions, 
if  $t_{all} < t_{max}$, then we use $t = t_{all}$. Otherwise, {\em at each pre-training iteration} we randomly select a time series $\pmb{s}$ of  $t = t_{max}$ consecutive transactions over the  sequence of all possible $t_{all}$ rows of that client. 
In \Cref{tab.dataset-ceco} we present results with $t_{max}$ ranging from 50  to 150.
This experiment is important to test whether a model dealing with time series can operate  in real life scenarios  where the actual row sequence length is variable and it can be very long ($t \geq 50$).

 We fine-tune the models  similarly to the  Fraud detection task (i.e., using a \texttt{[CLS]} token, etc.).
 Both in the fine-tuning and in the testing stage, the length $t$ of a transaction is variable. However, differently from the pre-training stage, if $t_{all} > t_{max}$, then we select the {\em last} $t_{max}$ transactions.

 \Cref{tab.dataset-ceco} shows that
 UniTTab   outperforms both TabBERT
 and LUNA,
 especially  when using very long sequences ($t_{max} = 150$).
The bottom of that table shows the results of VAR, XGBoost and CatBoost, with the usual dataset and task specific  features and hyperparameters selection for XGBoost and CatBoost. Additionally,
we also report the result obtained by 
\citet{Random-Forest} 
using Random Forests with 18 features, some of which are computed extracting aggregated information from the time series, while others 
are socio-demographic information about the client  which all the other methods  do {\em not} use. 
The results in \Cref{tab.dataset-ceco}  show that also in this small dataset UniTTab outperforms the standard Machine Learning methods and, similarly to the other tasks, the benefit is larger with longer time series.

\begin{table*}[h]
\caption{Loan default prediction task: average and standard deviation results obtained with 5 random seeds. 
\textsuperscript{\dag}~Our reproduction.
\textsuperscript{\ddag}~Results  reported in the corresponding  paper.
}
\label{tab.dataset-ceco}
\begin{center}
\footnotesize{
\begin{tabular}{@{}llllll@{}}
\toprule
Pre-training  & Model &   F1 score \\ 
 ($t_{max}$)  &  &     \\ \midrule

\multirow{3}{*}{50} 
& TabBERT  \textsuperscript{\dag}      &  $0.611 (\pm 0.032)$  \\ 
& LUNA  \textsuperscript{\dag}      &  $0.604 (\pm 0.048)$ \\ 
&  UniTTab (ours)                      &  ${\bf 0.619} (\pm 0.011)$ \\ \midrule
\multirow{3}{*}{100} 
& TabBERT  \textsuperscript{\dag}      & $0.636 (\pm 0.024)$   \\ 
& LUNA  \textsuperscript{\dag}      &  $0.624 (\pm 0.075)$ \\
&  UniTTab (ours)                      & ${\bf 0.654} (\pm 0.032)$  \\ \midrule
\multirow{3}{*}{150} 
& TabBERT  \textsuperscript{\dag}      & $0.620 (\pm 0.024)$    \\ 
& LUNA  \textsuperscript{\dag}      &  $0.637 (\pm 0.043)$  \\
&  UniTTab (ours)              & ${\bf 0.673} (\pm 0.038)$    \\  \midrule
 & Random Forest \citep{Random-Forest} \textsuperscript{\ddag} & 0.2667   \\ 
&  XGBoost \textsuperscript{\dag} &    $0.608 (\pm 0.079)$  \\ 
 &  CatBoost \textsuperscript{\dag} &    $0.527 (\pm 0.065)$   \\ 
 
 &  VAR \textsuperscript{\dag} &  $0.474 (\pm 0.007)$       \\ 
 \bottomrule
\end{tabular}
}
\end{center}
\end{table*}

\begin{table*}[h]
\caption{Age prediction task on the Age2 dataset.
\textsuperscript{\dag}~Our reproduction.
}
\label{tab.Age2}
\begin{center}
\small{
\begin{tabular}{@{}lll@{}}
\toprule
Downstream   & Model &   F1 score  \\ 
task training &  &     \\ \midrule
\multirow{3}{*}{Fine-tuning} 
& TabBERT \textsuperscript{\dag}     & 0.662  \\ 
&  UniTTab (ours) &    {\bf 0.678}  \\ \midrule
\multirow{3}{*}{LSTM} 
& TabBERT \ddag \cite{TabBERT}    & 0.645  \\ 
& LUNA \ddag \cite{LUNA}    & 0.648  \\ 
&  UniTTab (ours) &    {\bf 0.664}    \\ \midrule
Training from &  XGBoost \textsuperscript{\dag} &    0.542    \\ 
scratch &  CatBoost \textsuperscript{\dag} &    0.622   \\
&  VAR \textsuperscript{\dag} &    0.354    \\\bottomrule
\end{tabular}
}
\end{center}
\end{table*}

\begin{table*}[h]
\caption{Churn  prediction task on the RBAT Dataset. 
\textsuperscript{\dag}~Our reproduction.
}
\label{tab.churn}
\begin{center}
\begin{tabular}{@{}lll@{}}
\toprule
 Pre-training  & Model &   F1 score  \\ 
 ($t_{max}$)  &  &     \\ \midrule
\multirow{3}{*}{150} 
& TabBERT   \textsuperscript{\dag} &  0.526  \\ 
& TabBERT + VRT  (ours) &  0.536   \\ 
&  UniTTab  (ours) &    {\bf 0.604}  \\ \midrule
 &  XGBoost \textsuperscript{\dag} &    0.485  \\ 
 &  CatBoost \textsuperscript{\dag} &    0.483   \\
 &  VAR \textsuperscript{\dag} &    0.472    \\ 
 
 \bottomrule
\end{tabular}
\end{center}
\end{table*}

{\bf Age  prediction task.}
These experiments are based on the Age2 Dataset and its associated  Age prediction downstream task, both described in \Cref{sec:setting}. Similarly to the PKDD’99 Financial Dataset, Age2 is 
composed of the bank transaction history of different bank customers. Each row (transaction) is composed of $k = 3$ fields ($5$ after splitting the time stamp in $3$  fields).
For data augmentation, we follow the same protocol adopted when pre-training on the PKDD’99 Financial Dataset, i.e., we use time series of variable length $t \leq t_{max}$, extracted at random at each iteration from the whole history of each bank account. We  use $t_{max} = 50$.
Similarly to the Loan default prediction task, 
during both fine-tuning and testing, we use time series with variable length 
 $t$ and, if $t_{all} > t_{max}$, then we select the {\em last} $t_{max}$ transactions.

{\bf Churn  prediction task.}
In this last battery of experiments, we use the large RBAT Dataset 
and its associated downstream task (both described in  \Cref{sec:setting}).
Similarly to the PKDD’99 Financial Dataset and the Age2 Dataset, also this dataset is composed of the bank transaction history of different real users. However,  RBAT is much more complex than most public datasets, and each transaction (row) of its time series is associated with a specific type. To be precise, there are $n = 3$
types of mutually exclusive transactions (see \Cref{sec.Preliminaries}):
(1) generic transactions, with  $k_g = 5$ fields, (2) POS transactions, with $k_p = 8$  fields and (3) ATM transactions, with $k_a = 7$  fields. Since the generic transaction fields are shared also by the other two types of transactions, the total number of different fields is 10, which become 12 when the time stamp is split in 3  different fields (day, month, year).

For data augmentation, we follow the same protocol adopted when pre-training on both the PKDD’99 Financial Dataset and the Age2 Dataset, i.e., we use time series of variable length $t \leq t_{max}$, extracted at random at each iteration from the whole history of each account. In this case, we  use $t_{max} = 150$.
Since the baseline
TabBERT cannot deal with different types of transactions and needs a fixed number of fields for each row, when training TabBERT we concatenate all the 
 $k = 10$  fields of all the row types. In a given row $\pmb{r}$, the field values of those attributes which are not included in the type of $\pmb{r}$, are  represented using a \texttt{[MISSING]} token.
Note that this solution is computationally  more expensive, especially in the Field Transformer, where the computation costs grow quadratically with $k$, and the computational benefit of our proposal is larger when $n$ is bigger.
For ablation reasons (see \Cref{sec.Ablation}), we also train a modified version of TabBERT in which we introduce the Row-type dependent embedding (\Cref{eq.row-specific-transform}) {\em without any other changes}.
This baseline corresponds to the baseline used in \Cref{tab.ablation} (first row), on top of which we add the Row-type dependent embedding described in \Cref{sec.Method}, thus we call it  ``TabBERT + Variable Row Types'' (TabBERT + VRT).

We define the Churn  prediction task in \Cref{sec:setting}.
Similarly to the Loan default prediction task and the Age prediction task, 
during both fine-tuning and testing, we use time series with variable length 
 $t$ and, if $t_{all} > t_{max}$, then we select the {\em last} $t_{max}$ transactions (which are the closest to a possible bank account closure).
Since this dataset is much larger than the others, 
for computational reasons we have not included LUNA in this comparison. However, similarly to the  other tasks, we also used VAR, XGBoost and CatBoost with the usual dataset and task specific  features and hyperparameter selection for the last two.

\Cref{tab.churn} shows that TabBERT + VRT significantly improves the baseline TabBERT, showing that   the Row-type dependent embedding  has an accuracy benefit on the downstream task which goes beyond its computational advantages. Moreover,  also in this task our full method (UniTTab)  outperforms TabBERT, VAR, XGBoost and CatBoost with a large margin.

\subsection{Effect of pre-training}
\label{sec.Pre-Training}

One of the main advantages of using Deep Learning methods over more traditional Machine Learning approaches, is the possibility to pre-train a large network using self-supervision and a large unsupervised dataset, and then fine-tune the same network on the available supervised data of a downstream task. 
For instance, in case of UniTTab, we first pre-train the network using a Masked Token pretext task (\Cref{sec.Method}) using all the available training data. These data do not need to be annotated, since predicting a masked token is a self-supervised task. Then, we use downstream task-specific annotated training  data (which are usually much sparser than the unsupervised data) to fine-tune the network. This is not possible with techniques like VAR, Gradient Boosted Decision Trees or Random Forests, which need labeled data and thus cannot exploit the knowledge contained in the unlabeled samples.

In order to quantify the contribution of the pre-training phase, and to show that this is useful also when the unlabeled dataset is not huge,  we use  the two smallest datasets, i.e., the Pollution Dataset and the   PKDD’99 Financial Dataset,
 and we pre-train the models with different portions of the pre-traing dataset. Specifically, in both \Cref{tab.pollution-different-pre-training}
and \Cref{tab.dataset-ceco-different-pre-training}
we indicate the fraction of the pre-training dataset used for each experiment, where zero corresponds to training the models from scratch directly on the (labeled) downstream task data. The results in these tables show that both TabBERT and UniTTab significanltly benefit from the pre-training phase, despite both datasets have a small-medium size, and even when only a small portion of the unlabeled data (e.g., 0.25) is used for pre-training. Moreover, \Cref{tab.pollution-different-pre-training} shows that UniTTab can drastically outperform the non Deep Learning based methods (VAR,  XGBoost and CatBoost) even with no pre-training.
Conversely, in the smallest dataset (PKDD’99), XGBoost beats all the other methods with a large margin when no pre-training is used (\Cref{tab.dataset-ceco-different-pre-training}).
Note that this dataset is very small, with only 478 labeled samples (\Cref{sec:Dataset-statistics}), and training from scratch networks with 100M parameters (\Cref{sec:hyperparameters}) with these scarce data is very difficult. However, when pre-training is used, UniTTab gets a significantly higher F1 score than XGBoost.

\begin{table}[h]
\caption{Pollution prediction task (our partition): impact of different portions of the pre-training dataset.
\textsuperscript{\dag}~Our reproduction.}
\label{tab.pollution-different-pre-training}
\begin{tabular}{@{}lll@{}}
\toprule
Pre-training portion & Model &   RMSE  \\ \midrule
\multirow{2}{*}{0} 
& TabBERT  \textsuperscript{\dag}    & 33.37  \\
&  UniTTab (ours) &  {\bf 27.10}   \\ 
&  XGBoost \textsuperscript{\dag} &    48.89    \\ 
&  CatBoost \textsuperscript{\dag} &    47.09 \\
&  VAR \textsuperscript{\dag} &    84.05    \\ 

\midrule
\multirow{2}{*}{0.25} 
& TabBERT  \textsuperscript{\dag}    & 30.78  \\
&  UniTTab (ours) &  {\bf 24.55}  \\ \midrule
\multirow{2}{*}{0.5} 
& TabBERT  \textsuperscript{\dag}    & 30.11  \\
&  UniTTab (ours) &  {\bf 23.73}  \\ \midrule
\multirow{2}{*}{0.75} 
& TabBERT  \textsuperscript{\dag}    & 29.51  \\
&  UniTTab (ours) &  {\bf 23.32}   \\ \midrule
\multirow{2}{*}{1} 
& TabBERT  \textsuperscript{\dag}    & 29.13  \\
&  UniTTab (ours) &   {\bf 23.29}   \\ \bottomrule
\end{tabular}
\end{table}

\begin{table*}[h]
\caption{Loan default prediction task: impact of different portions of the pre-training dataset.
We report average and standard deviation results obtained with 5 random seeds. For both  TabBERT
and UniTTab, we keep fixed $t_{max} = 150$.
\textsuperscript{\dag}~Our reproduction.
\textsuperscript{\ddag}~Results  reported in the corresponding  paper.
}
\label{tab.dataset-ceco-different-pre-training}
\begin{center}
\footnotesize{
\begin{tabular}{@{}lll@{}}
\toprule
Pre-training  & Model &   F1 score  \\ 
portion &  &   \\ \midrule
\multirow{5}{*}{0} 
& TabBERT  \textsuperscript{\dag}       &    0.526 $(\pm 0.009)$  \\ 
&  UniTTab (ours)                       &    0.548 $(\pm 0.015)$ \\
 & Random Forest \citep{Random-Forest} \textsuperscript{\ddag} & 0.2667    \\ 
&  XGBoost \textsuperscript{\dag} &    ${\bf 0.608} (\pm 0.079)$  \\ 
 &  CatBoost \textsuperscript{\dag} &    $0.527 (\pm 0.065)$    \\ 
 &  VAR \textsuperscript{\dag} &    $0.474 (\pm 0.007)$    \\ 
 \midrule
\multirow{2}{*}{0.25} 
& TabBERT  \textsuperscript{\dag}       &    0.586 $(\pm 0.023)$  \\ 
&  UniTTab (ours)                       &  {\bf  0.593} $(\pm 0.022)$  \\ \midrule
\multirow{2}{*}{0.5} 
& TabBERT  \textsuperscript{\dag}       &    0.577 $(\pm 0.022)$  \\ 
&  UniTTab (ours)               &   {\bf 0.607} $(\pm 0.029)$   \\   \midrule
\multirow{2}{*}{0.75} 
& TabBERT  \textsuperscript{\dag}       &  {\bf  0.628} $(\pm 0.019)$  \\ 
&  UniTTab (ours)               &    0.620 $(\pm 0.027)$   \\   \midrule
\multirow{2}{*}{1} 
& TabBERT  \textsuperscript{\dag}       &    $0.620 (\pm 0.024)$    \\ 
&  UniTTab (ours)               &    ${\bf 0.673} (\pm 0.038)$   \\   \bottomrule
\end{tabular}
}
\end{center}
\end{table*}

\section{Conclusions}
\label{sec.Conclusions}

We proposed  UniTTab, a hierarchical Transformer architecture which can uniformly process highly  heterogeneous  time series  of tabular data with
variable lengths, including categorical and numerical values, as well as rows with different internal structure and type. 
UniTTab can be pre-trained using a uniform Masked Token task (independently of the feature type input), and 
fine-tuned for different tasks. Our experiments show that the proposed method consistently  outperforms the state-of-the-art tabular time series approaches, usually with a large margin. 
We believe that our architecture and our unified Masked Token pre-training  can 
 pave the way to large scale pre-trained  foundation models in the domain of tabular data with heterogeneous attributes.

\backmatter


\section*{Declarations}

\begin{itemize}
\item Funding: Partial financial support was received from Prometeia Spa, Bologna, Italy, and from Prometeia Associazione, Bologna, Italy.
\item Conflict of interest/Competing interests: One of the authors, specifically Enver Sangineto, is member of the Editorial Board of the Machine Learning Journal.
\item Ethics approval: Not applicable. 
\item Consent to participate: Not applicable.
\item Consent for publication: Not applicable.
\item Availability of data and materials: All the dataset are public and available online, except our private RBAT Dataset, which, for  both privacy and commercial reasons,  cannot be released.
\item Code availability: The code is currently not publicly available. 
\item Authors' contributions: Simone Luetto and Fabrizio Garuti wrote the code and conducted all the experiments.  Enver Sangineto and Rita Cucchiara, jointly with Luetto and  Garuti, proposed the ideas the method is based on and conducted the research. Lorenzo Forni provided the funding  and contributed to the research with his expertise in financial topics. All authors wrote and reviewed the paper.
\end{itemize}

\bibliography{sn-bibliography}

\begin{appendices}

\section{Experimental setting}
\label{sec:setting}

In this section, we present in more detail the datasets and the associated downstream  tasks we used for our evaluations. Generally speaking, the five datasets we used are  very different from each other in terms of size, attributes,  complexity and downstream  tasks. 
Moreover, we use two main protocols to extract the time series and split the data in training and testing partitions, the first proposed in \citep{TabBERT}, and the second proposed in this paper.
Specifically, for both the Pollution Dataset and the Transaction Dataset we follow the protocol proposed in \citep{TabBERT}, in which time series are extracted from longer sequences using a  temporal sliding window (whose length corresponds to the time series length $t$) and a stride. 
These time series are then randomly split in training and testing sequences using a fixed training/testing ratio and used for the corresponding downstream task. 
Note that the {\em pre-training} data include the downstream task  testing
 samples, however, {\em no downstream task label is used in pre-training}, since this phase is completely unsupervised. Even if a clearer separation between the pre-training dataset and the downstream task testing dataset would be desirable, we followed the protocol proposed in \citep{TabBERT} to make possible a comparison with the methods that have been evaluated with those benchmarks using the same approach \citep{TabBERT,LUNA,TabAConvBERT}.
 On the other hand, for the other three datasets adopted in this paper, the PKDD’99 Financial Dataset, the Age2 Dataset and the RBAT Dataset (all bank transaction datasets),  we split the time series based on the bank client, which leads to a more realistic scenario and a sharper training-testing separation. Specifically, we create two data partitions (training and testing) based on the bank customer identity, and in each partition  we collect the entire transaction history of the corresponding clients. Therefore, the history of a given customer can either belong (entirely) to the training data or (entirely) to the testing data. 
 The training split is used both in the pre-training phase and in the downstream task training phase, while testing data are observed by the models  only at the time of evaluating the downstream task.
 Finally, as explained in \Cref{sec.Main-Results}, for each dataset we define a $t_{max}$ length and we extract time series with variable  length  $t$, where $t \leq t_{max}$.
 We provide below more details on each dataset/task, while in \Cref{sec:Dataset-statistics} we show the statistics of each dataset.

{\bf The Pollution Dataset} 
 \citep{PollutionDataset} 
is   a public
UCI dataset  based on pollution  data collected  from 12 monitoring sites. Every row is composed of $k=11$  fields ($k=14$ using our time stamp representation with 4 fields) and 
it was adopted in \citep{TabBERT} to extract 
 time series using the concatenation of $t$ 
time-dependent, consecutive
rows, obtained with a $t$ long sliding window  and a stride of 5 (see below).
Following \citep{TabBERT}, in \Cref{tab.pollution-TabBERT-split,tab.pollution-our-split}  we  {\em pre-train}   all the models using 76K time series with length $t = 10$.

For evaluation, we adopt the Pollution prediction (supervised)  downstream task   \citep{TabBERT}, consisting in predicting the air  pollution  concentration. For this  task,
 \citet{TabBERT}
use 45K (supervised) time series samples for training and 15K (supervised) samples for testing, 
in which the time series are obtained using a sliding window and a stride of 5.
However, using this stride, there is an overlapping between adjacent sequences, and, since the training-testing splits of the downstream task are obtained by random sampling of the time series,  the two splits may have  a partial information leak. To avoid this,  we use a   stride of 10 (with no overlapping), which leads to a downstream task  training-testing partition different from the one used in \citep{TabBERT}, with 23K training and 7.6K testing sequences.
As mentioned in \Cref{sec.Results},  we call 
``our partition'' the  training-testing random splits obtained using a stride of 10, and  
``original partition'' the  training-testing random splits used in \citep{TabBERT} and based on a stride of 5 (see \Cref{sec:Dataset-statistics} for more details).

Finally, in \Cref{tab.pollution-t-50} we use time series with length $t = 50$ and a stride of 50,
which leads to $\sim$7.6K unsupervised time series used for pre-training,  
$\sim$6.1K used for the supervised downstream task training and $\sim$1.5K testing sequences.

{\bf The Transaction Dataset} \citep{TabBERT} is a synthetic dataset created with heuristic rules to generate realistic credit card transactions (see \Cref{fig.transaction-dataset}).
The dataset is composed of
 24M transactions from 20,000
users. There are $k = 10$ attributes, which become $k = 13$ when the time stamp is split in 4 different fields (\Cref{sec.Method}).
Following \citep{TabBERT}, a time series is composed of $t = 10$ transactions (i.e., rows), 
and, for pre-training,  the  stride of the sliding window is 5,
which leads to 
a total number of  4.87M pre-training time series. 

The associated Fraud detection downstream  task 
is a binary classification task based on the prediction of the   fraudulent label, associated with a subset of  samples. 
Following \citep{TabBERT}, we use 
1.9M  labeled samples (with $t=10$) for the supervised training stage. 
Note that, 
for the downstream task, \citet{TabBERT} 
extract time series 
using a stride of 10, thus,
differently from the Pollution prediction  downstream task, there is no 
overlapping between two adjacent time series and no information leak
 in the splits defined in \citep{TabBERT}.
However, since the positive and the negative classes are highly unbalanced, the positive samples are upsampled  (for more details, we refer to \citep{TabBERT}). The testing set is composed of 487K time series.

For computational reasons, in the ablation experiments presented in \Cref{sec.Ablation} (\Cref{tab.ablation,tab.masking,tab.L-ablation}) we use only 400K randomly selected time-series for pre-training, of which 360K are used for the downstream task training and 40K for testing. 
Finally, in \Cref{tab.Fraud-detection-t-50}, we use $t=50$ and a stride of 50, which leads to 
$\sim$475K unsupervised time series for pre-training, 
$\sim$428K supervised time series for the downstream task training and $\sim$47K supervised downstream task testing time series.

{\bf The PKDD’99 Financial Dataset}
 \citep{Berka-Dataset}
 is a 
collection of real anonymized financial data of a Czech bank. Besides other client specific information (which we do  not use), this dataset contains bank account transactions of 4,500 clients. 
We use $k =$ 6 fields to represent each transaction (row),
 which become $k = 8$ after splitting  the time stamp in 3 different fields.
On average, 
the total length of transaction history for a given customer is 232.

The associated Loan default prediction downstream task is a binary classification task consisting in 
 predicting  how likely a client is going to default the loan.
 For this  task,  we use 682 clients, jointly with their loan ground truth label, 
 which, following \citep{Random-Forest}, are 
 split in 478 clients  for training and 204 for testing, and we 
 report the average results obtained with 5 random splits. 
 Note that 
 the testing clients are not used for pre-training and 
 we removed (from the pre-training, the fine-tuning and the testing data) all those transactions directly related to the loan payment.
Moreover, both at fine-tuning and at testing time, we cut the transaction history of each client before the loan started.
 
{\bf The Age2 Dataset} \citep{Age2} is a public dataset composed  of real
bank transactions of different clients. Specifically, there are 43K  clients, which we split in 
39K for training (used both in the pre-training stage and in the downstream task training phase) and  4K for testing (used only for evaluation).
The average length of the transaction history of each client is
$\sim$84 rows. As shown in \Cref{sec.Main-Results}, time series of variable length are extracted from each client with a maximum length of $t_{max} = 50$. 

The downstream task  requires the model to predict the age  of the client given a time series.
Specifically, we formulate this task as a binary classification task, where the models should
 predict whether the customer is over 30 years old.

{\bf The RBAT Dataset}
is a proprietary dataset  provided by an international bank\footnote{For  both privacy and commercial reasons, this dataset cannot be released.}, which is composed of several hundred  thousands real bank account transactions of private clients. From this datasets, we 
have randomly selected 100K bank accounts, corresponding to about 32.5M transactions (i.e., rows), which we used for both pre-training and fine-tuning on the donwstream task, and another  20K accounts used only for the downstream task testing.
Overall, the time series transactions   span about 2 years, from 2021 to 2022. For a given account, the average transaction history length is 325. 
There are  $n = 3$ types of transactions (see \Cref{sec.Main-Results}), and,
in the future, we plan to extend our experiments with larger portions of this dataset, including other types of transactions ($n > 3$).

 The Churn  prediction  downstream task consists in predicting a bank account closure  
after a period of one
 month from the last transaction considered. 
 In more detail, at inference time,
 given a time series $\pmb{s}$, extracted from a given bank account,  
 the model should predict a possible  closure of that account which can happen 
  any day of the month following the last transaction  contained in $\pmb{s}$
  (which is a supervised information provided at fine-tuning time).

\section{Dataset statistics}
\label{sec:Dataset-statistics}

In  \Cref{tab.Dataset-statistics-1,tab.Dataset-statistics-2} we report the main characteristics of the datasets used in our experiments, including our RBAT Dataset. Specifically, both
``Pre-training samples'' and ``Downstream training samples''
refer to the original number of time series samples {\em before} any upsampling or data-augmentation process.
The reported number of attributes include the time stamp counted as a single field.
For the  
Pollution Dataset,  we also report  the statistics of our partition, obtained using a non-overlapping stride which guarantees that there is  no information leak between the training and the testing split (\Cref{sec:setting}). 
For both the Pollution and Transaction datasets, we also report
the statistics corresponding to time series of length $t = 50$, used in \Cref{tab.pollution-t-50} and \Cref{tab.Fraud-detection-t-50}, respectively. Finally, the last column of \Cref{tab.Dataset-statistics-1} refers to the random subset of the Transaction Dataset used in the ablation experiments of \Cref{sec.Ablation}.

\begin{table*}[h]
\caption{Dataset statistics: Pollution Dataset and Transaction Dataset.} 
\label{tab.Dataset-statistics-1}
\begin{center}
\footnotesize{
\begin{tabular}{@{}lllllll@{}}
\toprule
     & \multicolumn{3}{c}{Pollution Dataset} & \multicolumn{3}{c}{Transaction Dataset }  \\ 
           & our partition      & original partition    & $t=50$       & original   & $t=50$ & ablation    \\
\midrule
Total dataset rows           & 382,168      & 382,168     & 382,168       & 24,386,900  & 24,386,900 & 4,000,000    \\
Number of attributes ($k$)   & 11           & 11          & 11            & 10          & 10         & 10             \\
Number of categorical fields & 1            & 1           & 1             & 8           & 8          & 8              \\
Number of numerical fields   & 10           & 10          & 10            & 2           & 2          & 2              \\
Time series length ($t$)     & 10           & 10          & 50            & 10          & 50         & 10               \\
Pre-training samples         & 76,414       & 76,414      & 7,638         & 4,874,597   &  475,066   &  400,000         \\
Downstream training          & 22,927       & 45,850      & 6,114         & 1,950,224   &  427,559   &  360,000         \\
samples                      &          &    &         &             &            &                 \\
Testing samples              & 7,642        & 15,282      & 1,524         & 487,556     &   47,507   &   40,000        \\ 
 \bottomrule    
\end{tabular}
}
\end{center}
\end{table*}

\begin{table*}[h]
\caption{Dataset statistics: PKDD’99 Financial Dataset, Age2 Dataset and RBAT Dataset.} 
\label{tab.Dataset-statistics-2}
\begin{center}
\footnotesize{
\begin{tabular}{@{}llll@{}}
\toprule
     & PKDD’99 Financial   & Age2 Dataset & RBAT Dataset  \\ 
     & Dataset             &              &               \\ 
\midrule
Total dataset rows           & 1,042,740    & 3,652,757     & 39,040,010          \\
Number of attributes ($k$)   & 6            & 11          & 10                 \\
Number of categorical fields & 3            & 1           & 7                  \\
Number of numerical fields   & 3            & 10          & 3                 \\
Time series length ($t$)     & variable     & variable    & variable                   \\
Pre-training samples         & 4,500        & 38,961      & 100,000                \\
Downstream training          & 478         & 38,961      & 100,000                \\
samples                      &             &             &                 \\
Testing samples              & 204         & 4,328       &  20,000               \\   
\bottomrule    
\end{tabular}
}
\end{center}
\end{table*}

\section{Implementation details}
\label{sec:hyperparameters}

Following TabBERT, we use Positional Encoding only in the Sequence Transformer, being the attribute specific embeddings of the Field Transformer sufficient to the network to distinguish the specific field (intuitively, the attribute order in a row does not matter, once the network can distinguish an attribute from the others). 
\Cref{tab.Hyperparameter-values} shows all the hyperparameter values used in UniTTab, most of which are shared with TabBERT. 

We used Pytorch 1.11 and we trained all the models on 4 GPUs NVIDIA RTX A6000 (48G memory).
Note that  the model sizes reported in \Cref{tab.Hyperparameter-values} vary across datasets  almost only because of the difference in the size of the corresponding attribute vocabularies, which is an inherently task-dependent hyperparameter.

{\bf Regression loss.} In the ablation experiments of \Cref{tab.ablation}
(\Cref{sec.Ablation}) we have indicated with 
``Regression loss''
the  use of the MSE loss function for the numerical features, which is summed with the Cross Entropy loss used for the categorical attributes. We provide more details below.
When using ``Regression loss'', \Cref{eq.masked-token-loss} is replaced by the following objective function:

\begin{equation}
    \label{eq.regression-loss}
    \begin{split}
    \min_{\theta} 
    - \sum_{v \in \pmb{s}, Cat(v)} \mathbbm{1}_{Masked(v)}  \log p_{\theta} (quantize(v) | \pmb{s}') \\
    + \lambda \sum_{v \in \pmb{s}, \neg Cat(v)} \mathbbm{1}_{Masked(v)}  (\hat{v} -  f_{\theta} ( \pmb{s}'))^2. 
    \end{split}
\end{equation}

\noindent
Note that we do not use $\pmb{p}(v)$ because, in this ablation experiment, Label Smoothing is not used (\Cref{sec.Ablation}).
In \Cref{eq.regression-loss},  $Cat(v)$ is true when $v$ is a categorical attribute. In that case, the standard Cross Entropy loss is used, where the posterior $p_{\theta}(\cdot)$ is obtained with a softmax layer on top of the final MLP network (\Cref{sec.Preliminaries}). 
On the other hand, for numerical features ($\neg Cat(v)$), 
we use a linear regression layer (indicated by
$f_{\theta}(\cdot)$ in \Cref{eq.regression-loss}).
Moreover, $\hat{v}$ is the normalized $v$ value. In more detail, following \citep{TabBERT}, numerical feature values $v$  are pre-processed by applying a log-scale and a standardization using the mean and the standard deviation of the training sample values. 

The value of $\lambda$ is obtained using the following heuristic. 
We train the entire network for one epoch using $\lambda = 1$. Then, we use a few batches ($Batch$) to separately compute the average magnitude of the two losses:

\begin{equation}
    \label{eq.Loss-C}
     L_C = \sum_{\pmb{s} \in Batch}
    \sum_{v \in \pmb{s}, Cat(v)} \mathbbm{1}_{Masked(v)}  \log p_{\theta} (quantize(v) | \pmb{s}'),
\end{equation}

\begin{equation}
    \label{eq.Loss-R}
     L_R = \sum_{\pmb{s} \in Batch}    
     \sum_{v \in \pmb{s}, \neg Cat(v)} \mathbbm{1}_{Masked(v)}  (\hat{v} -  f_{\theta} ( \pmb{s}'))^2.
\end{equation}

\noindent
Finally, we impose that $L_C$ must be equal to $\lambda L_R$  and we solve for $\lambda$:

\begin{equation}
    L_C = \lambda L_R; \: \: \:  \: \: \: \lambda = \frac{L_C}{L_R}.
\end{equation}

Using this heuristic,
for the Pollution Dataset, we get $\lambda = 50$  and, for the Transaction Dataset, we get $\lambda = 25$.
As mentioned in \Cref{sec.Introduction}, the need to set the relative weighting factor ($\lambda$) for each dataset makes  difficult to simultaneously use heterogeneous loss functions. On the other hand, our proposed uniform  Masked Token pretext task (\Cref{eq.masked-token-loss}) 
besides being more effective from a performance point of view
(\Cref{sec.Ablation})
is also much more easy to use.

\begin{table*}[h]
\caption{UniTTab hyperparameter values.}
\label{tab.Hyperparameter-values}
\begin{center}
\begin{tabular}{@{}llllll@{}}
\toprule
                                  & Pollution             & Transaction      & PKDD’99        & RBAT      & Age2     \\ 
                                  & Dataset               & Dataset          & Financial      & Dataset   & Dataset  \\ 
                                  &                       &                  & Dataset        &            &   \\ 
\midrule 
Optimizer                                  & AdamW             & AdamW        & AdamW         & AdamW    & AdamW     \\
Learning rate                              & 5e-05             & 5e-05        & 5e-05         & 5e-05    & 5e-05     \\
Dropout                                    & 0.1               & 0.1          & 0.1           & 0.1      & 0.1       \\
Label Smoothing ($\epsilon$)               & 0.1               & 0.1          & 0.1           & 0.1      & 0.1       \\
Batch size                                 & 120               & 120          & 120           & 120      & 120       \\
Model size (parameters)                    & 137M              & 300M         & 100M          & 90M      & 44M      \\
Field Transformer layers                   & 1                 & 1            & 1             & 1        & 1         \\
Field Transformer heads                    & 8                 & 8            & 8             & 8        & 8         \\
Field Transformer                          &  72              &  72           &  72           &  72      &  72         \\
embedding size ($d$)                       &                   &              &               &          &           \\
Sequence Transformer                       & 12                & 12           & 12            & 12       & 12        \\
 layers                                    &                   &              &               &          &           \\
Sequence Transformer                       & 12                & 12           & 12            & 12       & 12        \\
 heads                                     &                   &              &               &          &           \\
Sequence Transformer                       &  1080            &  1080        &  1080          &   1080   &   1080        \\
embedding size ($m$)                       &                   &              &               &          &           \\

Pre-training epochs                        & 12                & 5            & 20            & 20       & 20         \\
Pre-training iterations                    & 7.6k              & 203k         & 0.76k         & 17k      & 21k      \\
Fine-tuning epochs                         & 10                & 10           & 30            & 30       & 20        \\
Fine-tuning iterations                     & 5.7k              & 485k         & 0.21k         & 30k      & 21k      \\
Total training time                        & 1.5 hours         & 4 days       & 15 minutes    & 8 hours  & 2 hours   \\ 
\bottomrule
\end{tabular}
\end{center}
\end{table*}




\end{appendices}


\end{document}